\documentclass[preprint,12pt,authoryear]{elsarticle}

\usepackage{amssymb}
\usepackage{amsmath}
\usepackage{amsfonts}
\usepackage{algorithmic}
\usepackage{algorithm}
\usepackage{array}
\usepackage[caption=false,font=normalsize,labelfont=sf,textfont=sf]{subfig}
\usepackage{textcomp}
\usepackage{stfloats}
\usepackage{url}
\usepackage{verbatim}
\usepackage{graphicx}
\usepackage{cite}
\usepackage{algorithm, algorithmic}
\usepackage{booktabs, multirow, makecell}
\usepackage{bm}
\usepackage{refcount}  
\usepackage{hyperref}  

\usepackage{pifont}
\usepackage[table]{xcolor}
\definecolor{gray}{rgb}{0.5,0.5,0.5}
\definecolor{Gray}{gray}{0.9}

\journal{Nuclear Physics B}

\begin{document}

\begin{frontmatter}



\title{End-to-End Spoken Grammatical Error Correction} 


\author{Mengjie Qian, Rao Ma, Stefano Bannò, Mark J.F. Gales, Kate M. Knill} 


\affiliation{organization={ALTA Institute, Department of Engineering},
            addressline={University of Cambridge}, 
            country={UK}}

\begin{abstract}
Grammatical Error Correction (GEC) and feedback play a vital role in supporting second language (L2) learners, educators, and examiners. 
While written GEC is well-established, spoken GEC (SGEC), aiming to provide feedback based on learners' speech, poses additional challenges due to disfluencies, transcription errors, and the lack of structured input. SGEC systems typically follow a cascaded pipeline consisting of Automatic Speech Recognition (ASR), disfluency detection, and GEC, making them vulnerable to error propagation across modules. 
This work examines an End-to-End (E2E) framework for SGEC and feedback generation, highlighting challenges and possible solutions when developing these systems. Cascaded, partial-cascaded and E2E architectures are compared, all built on the Whisper foundation model. 
A challenge for E2E systems is the scarcity of GEC labeled spoken data. To address this,  an automatic pseudo-labeling framework is examined, increasing the training data from 77 to over 2500 hours. To improve the accuracy of the SGEC system, additional contextual information, exploiting the ASR output, is investigated. 
Candidate feedback of their mistakes is an essential step to improving performance. In E2E systems the SGEC output must be compared with an estimate of the fluent transcription to obtain the feedback.
To improve the precision of this feedback, a novel reference alignment process is proposed which aims to remove hypothesised edits resulting from fluent transcription errors. Finally, these approaches are combined with an edit-confidence estimation approach, to exclude low-confidenceedits.
Experiments on the in-house Linguaskill (LNG) corpora and the publicly available Speak \& Improve (S\&I) corpus show that the proposed approaches significantly boost end-to-end SGEC performance. 
\end{abstract}



\begin{keyword}
Spoken Grammatical Error Correction \sep Pseudo-Labeling \sep Model Prompting \sep Confidence Estimation \sep Language Assessment and Feedback


\end{keyword}

\end{frontmatter}



\section{Introduction}
Grammatical error correction (GEC) plays a critical role in second language (L2) learning, especially in computer-assisted language learning (CALL). It helps learners improve their accuracy and fluency, and it supports teachers and examiners by providing consistent, objective feedback. While many tools now exist to correct written grammar, the ability to provide meaningful grammatical feedback on spoken language remains limited. As language learning shifts increasingly toward digital and automated platforms, especially in speaking skills, there is a clear need for systems that can accurately correct spoken grammar and offer useful feedback to learners.

GEC in written language is a well-established research area \citep{bryant2023grammatical,izumi2003automatic-full}, supported by a number of shared tasks such as CoNLL-2014 \citep{ng2014conll}, BEA-2019 \citep{bryant2019bea}, and MULTIGEC-2025 \citep{multigec2024}. These efforts have led to strong benchmark datasets and models capable of correcting a wide range of grammatical errors in text. However, written GEC operates under relatively clean conditions: input is well-structured, and errors are typically easier to detect due to the presence of punctuation and complete sentences. 
Spoken grammatical error correction (Spoken GEC or SGEC), by contrast, faces a different set of challenges. Spoken language is inherently noisy. It contains disfluencies (such as hesitations, repetitions, and false starts), incomplete or fragmented sentences, accented speech, and lacks punctuation and capitalization. These factors make detecting and correcting grammatical errors in speech significantly more difficult than in written form.

Traditionally, SGEC systems have followed a cascaded pipeline. This typically involves an automatic speech recognition (ASR) module to transcribe audio into text, followed by a disfluency detection (DD) module to generate fluent transcriptions, and finally a text-based GEC module to correct the grammar~\citep{lu2020spoken,lu2022assessing,banno2023b_slate_grammatical,banno2024back}. While effective to some extent, this design often suffers from error propagation between stages, limiting the system's overall performance. More recent work explores end-to-end (E2E) approaches using large speech foundation models such as Whisper~\citep{radford2023robust}, which have the potential to simplify the architecture and reduce compounded errors~\citep{banno2024towards}. However, training such models effectively remains difficult due to the limited availability of large, high-quality annotated spoken datasets. To address this, \citet{qian2025scaling} introduced an automatic pseudo-labeling process that scales the SGEC training data, improving the performance of an end-to-end SGEC model. Despite these gains, generating high-quality feedback remains a challenge. The recent release of the Speak \& Improve (S\&I) Corpus~\citep{knill2024speakimprovecorpus,knill2025introducing,qian2025speak}, the first publicly available speech dataset with grammar error annotations, marks a major step forward and is expected to drive further research and progress in this area.


Large language models (LLMs), including ChatGPT~\citep{ouyang2022training,achiam2023gpt}, LLaMA~\citep{touvron2023llama,touvron2023llama2,grattafiori2024llama3}, T5~\citep{raffel2020exploring}, Gemma~\citep{team2024gemma,team2024gemma2,team2025gemma3}, Mistral~\citep{jiang2023mistral7b}, and Qwen~\citep{bai2023qwen,yang2025qwen} have demonstrated strong capabilities in grammatical error correction~\citep{kobayashi2024large,davis2024prompting,katinskaia2024gpt} and error explanation~\citep{fei2023enhancing,song2024gee}. However, a well-documented limitation in learner-facing GEC applications is their tendency to over-correct, introducing unnecessary or stylistic changes rather than making minimal grammatical edits~\citep{fang2023chatgpt}.
This lack of minimal-edit precision reduces interpretability and reliability in feedback-oriented language learning contexts. Some methods have been proposed to attempt to mitigate this over-correction, for example \citet{staruch2025adapting}, based on adapting the LLMs for text-based grammatical correction. 


When applied to SGEC, LLMs are typically used as the second (final) stage in a pipeline, following an ASR module. 
Such hybrid ASR-to-LLM systems have been more commonly explored for ASR error correction~\citep{chen2023hyporadise}, where the goal is to recover the intended text from recognition errors rather than to correct grammatical mistakes in speech. Although similar in structure, ASR error correction and SGEC differ fundamentally: the former targets transcription inaccuracies, whereas the latter requires distinguishing between ASR errors, disfluencies, and genuine learner grammar issues.
LLMs are typically trained on well-structured and fluent text, and are not robust to disfluencies, fragmented syntax, or ASR errors, all of which are common in learner speech transcriptions. Similarly, the adaptation approaches for mitigating over-correction assume clean, well-formed written input \citep{staruch2025adapting}. These hybrid ASR-to-LLM pipelines inherit both transcription errors from the ASR and the over-correction tendencies of LLMs, making them suboptimal for learner feedback applications where precision and interpretability are crucial.

Emerging audio large language models (multimodal LLMs) such as AudioPaLM \citep{rubenstein2023audiopalm}, Qwen-Audio \citep{chu2023qwenaudio,chu2024qwenaudio2}, SALMONN \citep{tang2024salmonn}, and SpeechGPT \citep{zhang2023speechgpt} have shown exciting potential on tasks like speech translation and multimodal generation. These models are designed for broad general-purpose understanding and generation across speech and text modalities, and are often trained with multitask or instruction-following objectives. However, their application to SGEC remains limited for several reasons. These models lack the fine-grained control required for minimal-edit correction and explicit feedback generation, which are crucial in language learning contexts. Moreover, fine-tuning multimodal LLMs for SGEC is non-trivial: it often requires task-specific adaptation strategies, large amounts of annotated learner speech data, and careful handling of disfluent and accented input. While these limitations may be addressed as the field matures, they pose practical challenges for current SGEC tasks.
In contrast, Whisper~\citep{radford2023robust} is a speech foundation model primarily trained for ASR and speech translation tasks. Its training is focused on robust transcription from audio to text, which closely aligns with the core structure of SGEC, converting spoken learner input into grammatically correct transcriptions. This alignment makes Whisper more amenable to SGEC fine-tuning: it supports straightforward adaptation to the task, handles disfluencies well, and offers more control over output form.
For these reasons, this work focuses on Whisper-based E2E models for SGEC. 
Compared to general-purpose multimodal LLMs, Whisper's task-specific training and architecture make it a more practical and effective foundation for learning-focused grammar error correction and feedback.

In addition to producing grammatically corrected transcripts, SGEC systems must also provide useful feedback that helps learners understand and learn from their mistakes. Feedback is crucial in language learning, offering learners actionable insights into what went wrong, where the error occurred, and how to correct it. Despite its importance, current SGEC systems have limited capabilities in feedback generation, and research in this area remains sparse compared to the well-studied domain of written GEC. One early effort by \citet{lee2014grammatical} explored a statistical approach to error detection and feedback generation in spoken language. 
In \citet{lu2022assessing}, feedback edits are generated by comparing the inputs and outputs of the final GEC component of a cascaded SGEC pipeline (ASR$\rightarrow$DD$\rightarrow$GEC) and converting them to edit tags using ERRANT \citep{bryant2017automatic}.
More recently, \citet{banno2024towards} investigated feedback generation using an E2E Whisper-based model, but reported limited success in improving feedback quality. 
Notably, the same limitations that limit the use of LLMs and multimodal LLMs for SGEC, such as overcorrection, lack of edit transparency, and poor robustness to learner speech patterns, also hinder their effectiveness in feedback generation. These issues make it difficult to produce clear and interpretable feedback that aligns with pedagogical goals. As a result, grammatical feedback for spoken language remains largely unexplored, highlighting the need for new approaches tailored to the unique characteristics of learner speech.

This study extends previous work on end-to-end SGEC by proposing new methods to improve both grammatical error correction and feedback generation. 
Our focus is mainly on end-to-end SGEC systems that map learner speech directly to grammatically corrected transcriptions, rather than hybrid ASR-to-LLM pipelines. 
For completeness, we conducted preliminary experiments comparing the BERT-based GEC with GPT-4o in the GEC module of the partial-cascaded SGEC system, but found that GPT-4o underperformed, likely due to its overcorrection behavior and lack of adaptation to spoken input. Therefore, the BERT-based GEC is retained as a strong and consistent baseline for comparison with the Whisper-based end-to-end SGEC models evaluated in this work. Following \citet{lu2022assessing} and \citet{banno2024towards}, ``feedback'' in this work refers to edit-based grammatical feedback extracted from differences between fluent and corrected transcriptions.
Specifically, the contributions of this work are:
\begin{itemize}
    \item An updated and expanded evaluation of SGEC architectures (cascaded, partial-cascaded, and end-to-end) first introduced in \citet{banno2024towards}. While the prior study focused on small-scale models and single-domain data, this work adopts the architectures of Whisper small.en and large-v2, incorporates large-scale pseudo-labeled training data, and conducts comprehensive feedback-oriented evaluation on both proprietary and public benchmarks, providing the most comprehensive comparison of Whisper-based SGEC systems to date.
    \item A refined large-scale pseudo-labeling pipeline, building on the framework of \citet{qian2025scaling}, is enhanced with Whisper \texttt{large-v2} across all stages to improve transcription quality, consistency, and scalability. The resulting process produces over 2,500 hours of aligned speech–text data for SGEC, providing a robust and unified foundation for subsequent analyses.
     \item An extended prompt-conditioned training and inference strategy, originally explored in \citet{qian2025scaling}, is integrated into the expanded SGEC framework to enhance contextual grammatical correction and enable systematic evaluation together with the newly proposed reference alignment and edit-confidence methods.
    \item A novel reference alignment technique that improves the quality of training labels and the alignment between grammatically corrected outputs and fluent target transcriptions, enhancing feedback quality. This method enables, for the first time, an end-to-end SGEC model to surpass the partial-cascaded baseline in feedback generation, effectively addressing a long-standing limitation observed in earlier work~\citep{banno2024towards, qian2025scaling}.
    \item A new framework to estimate confidence for grammatical correction edits from the fluent and GEC transcriptions, and to filter out low confidence edits during evaluation. This two-sided word-to-edit aggregation provides more stable and interpretable feedback evaluation and complements reference alignment for further precision gains.
\end{itemize}

Although this work focuses on SGEC, the proposed techniques are not limited to this task. More broadly, they are applicable to cascaded spoken language processing systems in which ASR output is subsequently refined by text-based models, such as disfluency removal, ASR post-editing, and inverse text normalization.
The rest of this work is structured as follows: Section~\ref{sec:gec_systems} introduces the GEC system architectures considered in this work. Section~\ref{sec:methods} details the proposed methods, Section~\ref{sec:exp_setup} outlines the experimental setup, and Section~\ref{sec:results} presents results and key findings. Finally, Section~\ref{sec:conclusion} concludes with a summary and future directions.


\section{Spoken GEC System}
\label{sec:gec_systems}
\begin{figure}[!htbp]
    \centering
    \includegraphics[width=0.8\linewidth]{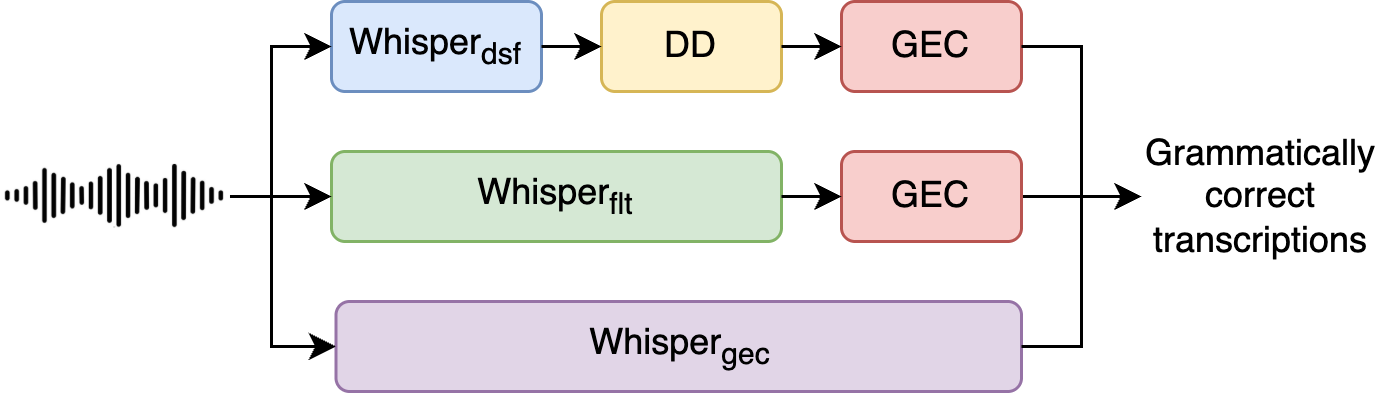}
    \caption{Illustration of the cascaded, partial-cascaded , and end-to-end SGEC systems.}
    \label{fig:sgec}
\end{figure}

Recent advances in large-scale speech foundation models, particularly Whisper \citep{radford2023robust}, have opened new possibilities for end-to-end spoken language processing. Whisper is trained on over 680,000 hours of multilingual audio with a multi-task learning objective, enabling it to generalize effectively across a wide range of tasks, including low-resource ASR \citep{qian2024learn, timmel2024fine}, speech translation \citep{ma2025cross}, and spoken language understanding \citep{ma2024investigating, goron2024improving}.
In this work, Whisper serves as the backbone for all three SGEC architectures explored: cascaded, partial-cascaded, and end-to-end (Figure~\ref{fig:sgec}). 
These architectures build upon the general framework of \citet{banno2024towards} but are extended to include both Whisper small.en and large-v2 backbones and trained on large-scale pseudo-labeled data. This enhanced setup establishes a unified and robust baseline for evaluating the newly introduced methods (Sections~\ref{sec:align_gec} and \ref{sec:confidence_desc}), ensuring that performance gains stem from methodological improvements rather than architectural or data differences.

\subsection{Cascaded System}
\label{sec:cascaded}
The cascaded SGEC system follows a modular structure composed of three sequential components: automatic speech recognition (ASR), disfluency detection (DD), and grammatical error correction (GEC), as shown in the upper part of Figure~\ref{fig:sgec}. The ASR module first transcribes learners' speech into text. Next, the DD module processes the transcript to detect and remove disfluencies such as hesitations, repetitions, and false starts, resulting in a more fluent version of the transcription. Finally, the GEC module identifies and corrects grammatical errors in the fluent text, producing grammatically corrected transcriptions.
This modular design benefits from the availability of task-specific resources (i.e. ASR training data, disfluency-annotated corpora, and large-scale written GEC corpora), making it a practical solution, especially in scenarios with limited annotated spoken GEC data. However, this approach is prone to error propagation, where mistakes in earlier stages (e.g. ASR errors) adversely affect the performance of downstream modules, ultimately reducing the quality of the final output. Moreover, because DD and GEC only receive text rather than the original audio, valuable acoustic information is lost, limiting the system's ability to make informed corrections. These limitations motivate the development of an end-to-end approach that can process audio directly and perform SGEC in a unified manner.

\subsection{Partial-cascaded System}
\label{sec:semi}
To reduce modular complexity and mitigate error propagation, we adopt a partial-cascaded system that integrates the ASR and DD components into a single model. 
The system consists of two stages, as illustrated in the middle part of Figure~\ref{fig:sgec}. First, a fluent speech transcription model, denoted as Whisper$_{\text{flt}}$, converts learner speech into a disfluency-free transcription. Second, a text-based GEC model processes this fluent transcription and generates the final grammatically corrected output. Compared with the fully cascaded architecture, this design removes the explicit DD module while retaining a separate GEC stage, thereby reducing error propagation and simplifying the overall pipeline.
Whisper$_{\text{flt}}$ differs from the standard Whisper in its training objective. When Whisper is fine-tuned for a standard ASR task (denoted Whisper$_{\text{dsf}}$ in this work), it is trained using verbatim transcriptions and aims to reproduce all spoken content that includes disfluencies. Whisper$_{\text{flt}}$ fine-tunes the standard Whisper model using fluent reference transcriptions as training targets. As a result, it learns to suppress disfluent content in the input audio and directly generates fluent transcriptions that more closely resemble edited written text. 
When combined with a GEC model, this setup strikes a balance between integration and interpretability, and serves as a strong baseline for our experiments.

\subsection{End-to-end System}
\label{sec:sgec_e2e}
End-to-end SGEC systems aim to directly map learner speech to grammatically corrected transcriptions using a single model (purple block in Figure~\ref{fig:sgec}). This approach simplifies the traditional multi-stage pipeline by removing intermediate modules such as ASR and disfluency detection.
In this work, the term \textit{end-to-end} refers specifically to the SGEC model, which directly maps learner speech to grammatically corrected transcriptions. Learner feedback is subsequently derived by comparing the corrected transcription with a fluent transcription and therefore additionally requires a fluent transcription generated by Whisper$_{\mathrm{flt}}$.
While our prior work~\citep{banno2024towards} introduced the general structure of an end-to-end SGEC, it showed limited success due to challenges like data scarcity.
We build on the E2E framework but make several key contributions that address those limitations in \citet{banno2024towards}.
By integrating large-scale pseudo-labeled data, leveraging prompt-based input conditioning, aligning references for more accurate supervision, and filtering out low confidence edits during evaluation, we substantially improve both grammatical correction and feedback quality. These enhancements allow the E2E system to not only match but in many cases surpass cascaded and partial-cascaded  baselines in both transcription accuracy and feedback effectiveness.

\section{Proposed Methods}
\label{sec:methods}
\begin{figure}[!t]
    \centering
    \includegraphics[trim=0 2mm 0 10mm, width=0.85\linewidth]{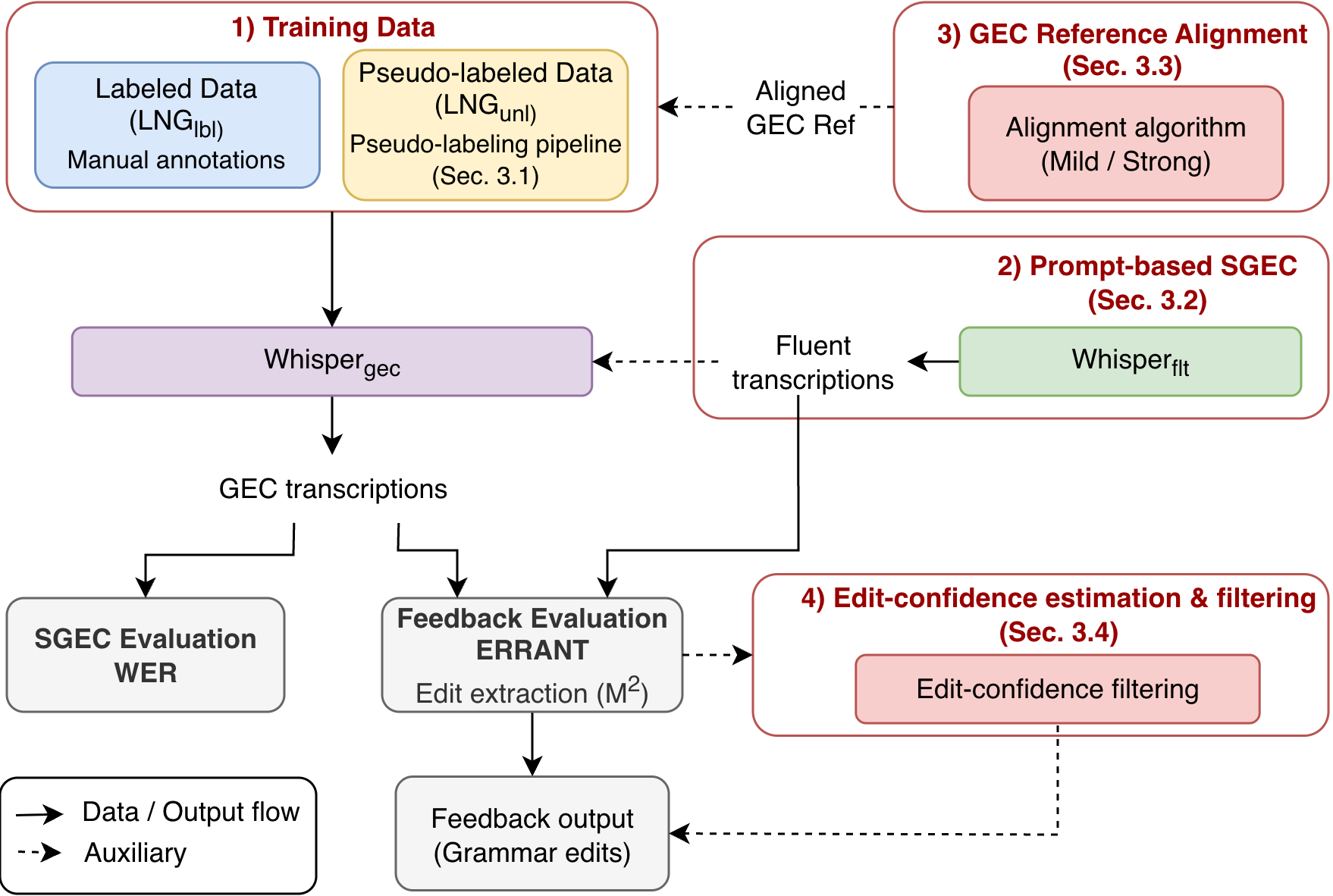}
    \caption{Overview of the proposed end-to-end SGEC framework with SGEC and feedback evaluation.}
    \label{fig:overall_framework}
\end{figure}

To address the limitations of existing SGEC systems, particularly in data availability and feedback quality, this work introduces four complementary techniques to enhance end-to-end SGEC performance. These are: (1) a large-scale automatic pseudo-labeling pipeline to automatically generate training data from unlabeled learner speech, (2) prompt-conditioned training and inference using fluent transcriptions to provide contextual guidance, (3) a novel GEC reference alignment strategy that refines training targets to better match the model's output structure and improve feedback generation quality, and (4) a new edit-level confidence estimation and filtering framework. The overall framework is shown in Figure~\ref{fig:overall_framework}.

\subsection{Automatic Pseudo-labeling}
\label{sec:pseudo-label}
End-to-end SGEC models require large quantities of annotated training data, but high-quality spoken GEC datasets remain scarce. In contrast, unlabeled audio recordings from language learners are widely available. To bridge this gap, we propose a fully automated pseudo-labeling pipeline that generates GEC-style annotations from raw audio, significantly increasing the amount of training data available for E2E model development.

Building on the pseudo-labeling framework of \citet{qian2025scaling}, we reconstruct the entire pipeline using Whisper \texttt{large-v2} across all stages 
to compare the impact of the model size in the pseudo-labeling process. 
Unlike the previous pipeline, which used smaller Whisper models and mixed configurations, the revised pipeline produces a unified pseudo-labeled dataset comprising over 2,500 hours of aligned audio, fluent, and grammatically corrected transcriptions. This dataset, denoted as LNG$_{\text{unl}}$, is the largest SGEC pseudo-labeled corpus to date and serves as the foundation for training the proposed end-to-end SGEC models. The pseudo-labeling pipeline (Figure~\ref{fig:pseudo-label_pipeline}) consists of the following steps:
\begin{figure}[!t]
    \centering
    \includegraphics[trim=0 2mm 0 10mm, width=0.65\linewidth]{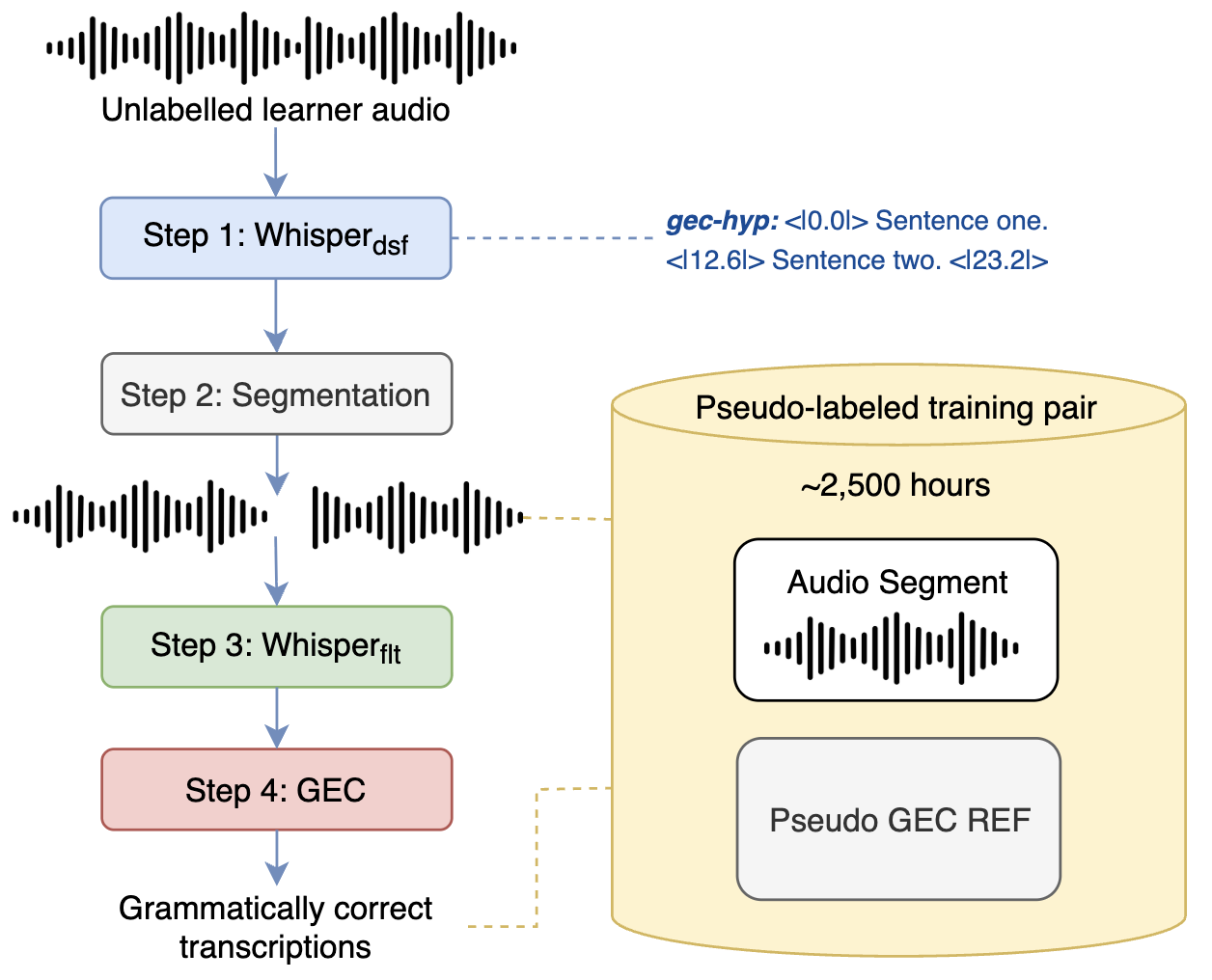}
    \caption{The SGEC pseudo-labeling pipeline.}
    \label{fig:pseudo-label_pipeline}
\end{figure}
\begin{itemize}
    \item Step 1: ASR with Disfluent Transcriptions. We fine-tune Whisper on 20 hours of manually segmented Linguaskill~\citep{ludlow2020official} data to generate disfluent transcriptions from raw audio. Fine-tuning incorporates segment-level timestamps derived from forced alignment using HTK~\citep{young2002htk} Hidden Markov Model (HMM)-Gaussian Mixture Model (GMM) MPE L2 English models, along with sentence-level truecasing, which is applied by capitalizing the first character of each sentence based on manual references.
    \item Step 2: Segmentation. Audio is segmented into phrase-level units using timestamp information and punctuation marks (full stops, question marks, and exclamation marks) detected in the output from Step 1, enabling more precise downstream alignment.
    \item Step 3: Generation of Fluent Transcriptions. Each audio segment is passed through Whisper$_\text{flt}$, a model fine-tuned for disfluency removal, to produce fluent transcriptions.
    \item Step 4: GEC Annotation. A text-based GEC system, previously trained on EFCAMDAT and BEA-2019 (see details on datasets in Section~\ref{sec:dataset}), then fine-tuned on Linguaskill, is used to correct fluent transcriptions, yielding phrase-level GEC targets.
\end{itemize}

Here, Step 3 and 4 adopt the best-performing SGEC pipeline established in prior work~\citep{banno2024towards} to ensure that the generated GEC transcriptions are of the highest possible quality. After Step 4, segments with a duration of 0s or longer than 30s are discarded, and the remaining segments from the same speaker are merged to a maximum of 30 seconds. No additional filtering is applied during pseudo-label generation. This results in a length distribution similar to the manually annotated LNG$_{\text{lbl}}$ dataset.
The final pseudo-labeled corpus contains approximately 2,500 hours of learner speech. Datasets generated using \texttt{small.en} in Steps 1 and 2 are denoted as LNG$_\text{unl}$ V1, while those generated using \texttt{large-v2} are denoted as LNG$_\text{unl}$ V2 (see Table~\ref{tab:data_stats} for statistics).



\subsection{Prompt-conditioned SGEC}
\label{sec:prompt}
\begin{figure}[!t]
    \centering
    \includegraphics[trim=0 2mm 0 5mm, width=0.7\linewidth]{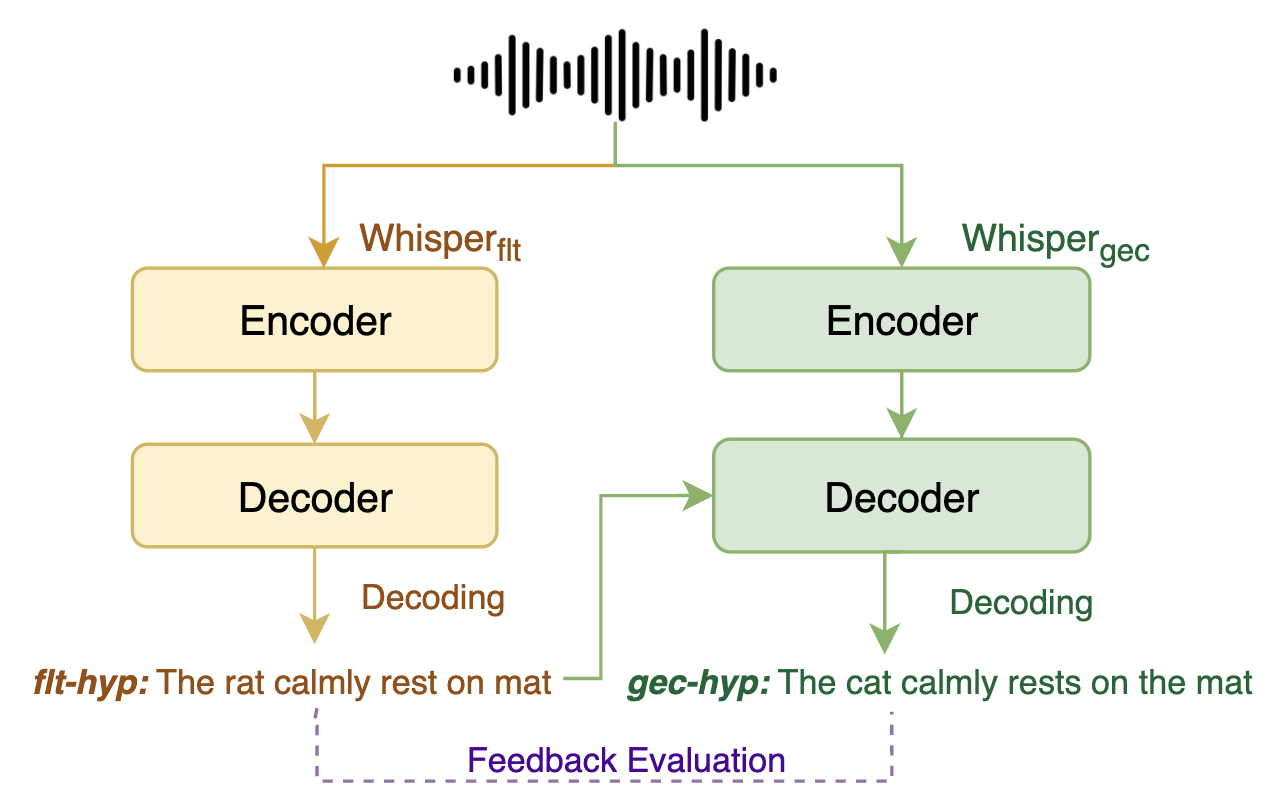}
    \caption{The end-to-end SGEC system prompting with additional ASR transcriptions. GEC reference: The cat calmly rested on the mat.}
    \label{fig:whisper-prompt}
\end{figure}
To improve the quality and informativeness of grammatical corrections, we propose a prompt-conditioned SGEC, where the model is conditioned on both the speech signal and a corresponding fluent transcription. The prompt is provided during both training and inference.
This approach extends prior work by introducing contextual prompting into the training of Whisper for SGEC, enabling the model to leverage an intermediate, cleaned transcript as a guide for grammatical correction. Unlike standard fine-tuning, where the model directly maps speech to corrected text, this method conditions the model on a fluent transcription generated by Whisper$_{\text{flt}}$ (Figure~\ref{fig:whisper-prompt}). 
During training, each example consists of (a) the original speech input, (b) a fluent transcription generated by Whisper$_\text{flt}$ as the model prompt, and (c) the corresponding grammatically corrected output as reference.
The prompt is also provided during inference, supplying additional linguistic context for grammatical correction. Furthermore, the explicit fluent transcription forms the basis for extracting learner feedback as edits between the fluent and grammatically corrected transcriptions. This design allows the model to focus on grammatical refinement rather than disfluencies or recognition errors. As shown in Section~\ref{sec:results_prompt}, the prompted framework consistently improves both SGEC performance and feedback quality.

\subsection{Aligned GEC Reference}
\label{sec:align_gec}
In SGEC, feedback quality is evaluated using metrics such as Precision, Recall, and F\textsubscript{0.5}, computed over the reference and hypothesis edit sets defined by the MaxMatch ($M^2$)~\citep{dahlmeier2012better} scoring algorithm (see Section~\ref{sec:metrics} for details). The reference edit set is obtained by comparing manual GEC transcriptions (\textit{gec-ref}) against manual fluent transcriptions (\textit{flt-ref}), while the hypothesis edit set is generated by aligning the system-produced GEC transcriptions (\textit{gec-hyp}) with their associated fluent transcriptions (\textit{flt-hyp}).
However, generating high-quality feedback in SGEC is particularly challenging due to disfluencies and errors introduced by ASR~\citep{banno2024towards,qian2025scaling}. These ASR artifacts can lead to spurious edits when comparing end-to-end SGEC (i.e. Whisper$_\text{gec}$) outputs (\textit{gec-hyp}) to fluent transcriptions produced by ASR (\textit{flt-hyp} from Whisper$_\text{flt}$), ultimately lowering the reliability of feedback evaluation. 

\begin{figure}[!t]
    \centering
    \includegraphics[trim=0 0mm 0 0mm, width=0.8\linewidth]{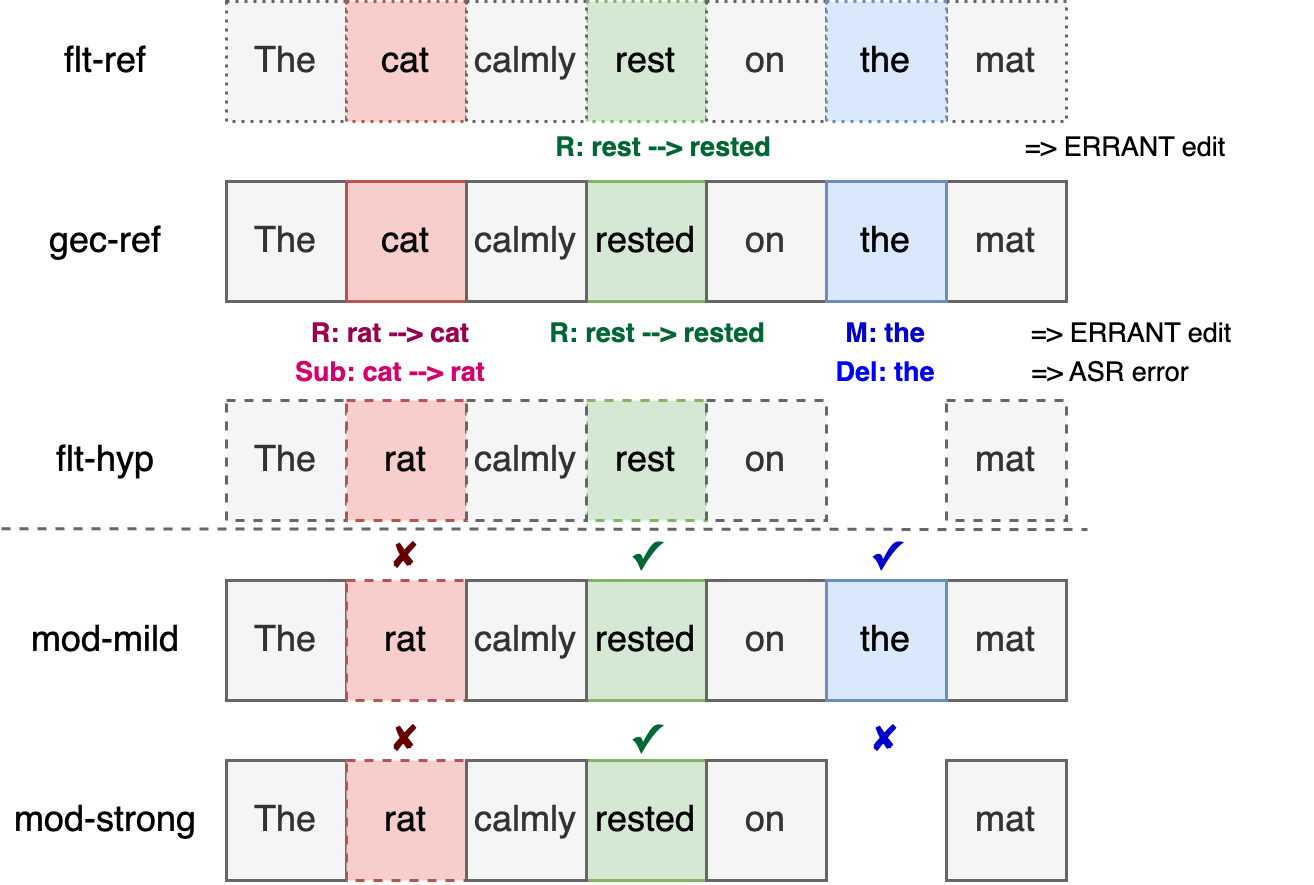}
    \caption{Illustration of the GEC reference modification process.}
    \label{fig:mod-gec}
\end{figure}

Figure~\ref{fig:mod-gec} illustrates the problem. Comparing the fluent reference (\textit{flt-ref}) and the GEC reference (\textit{gec-ref}), there is one genuine grammatical correction edit: ``rest'' is replaced with ``rested'', marked as an ``R:'' type edit in ERRANT, a widely adopted error classification tool for GEC. However, when comparing the fluent (FLT) hypothesis (\textit{flt-hyp}) to the same \textit{gec-ref}, three edits are detected. While one is valid (R: rest$\rightarrow$rested), the other two are false positives caused by ASR errors. Specifically, an ASR substitution error (``Sub: cat$\rightarrow$rat'') causes a replacement grammar edit (``R: rat$\rightarrow$cat''), while a deletion error (``Del: the'') leads to a missing type edit (``M: the''). These are not actual grammar errors but result from ASR noise, and including them in feedback evaluation or training may mislead both learners and the model.

To mitigate this issue, we propose a novel reference modification algorithm that adjusts the original GEC references to better align with the outputs of Whisper$_\text{flt}$ (i.e. \textit{flt-hyp}). This ensures that feedback generation focuses on true grammatical errors while minimizing the influence of ASR-related discrepancies.
The algorithm aligns three components:
\begin{enumerate}
    \item the FLT hypothesis (\textit{flt-hyp}): output of the Whisper$_\text{flt}$ model;
    \item the FLT reference (\textit{flt-ref}): human-annotated fluent transcription;
    \item the original GEC reference (\textit{gec-ref}): grammatically corrected transcription.
\end{enumerate}
The algorithm performs word-level alignments between (a) \textit{gec-ref} and \textit{flt-hyp}, and (b) \textit{gec-ref} and \textit{flt-ref}, focusing on the edits in the former that differ from those in the latter, then adjusts the \textit{gec-ref} using two strategies:
\begin{itemize}
    \item Mode mild: Keep words in the \textit{gec-ref} which correspond to insertions (``ins'' or ``U:'') or deletions (``del'' or ``M:'') of the \textit{flt-hyp}. Modify \textit{gec-ref} if the word corresponds to a substitution (``sub'' or ``R:''). This mode is more lenient and tackles more obvious ASR errors.
    \item Mode strong: Modify \textit{gec-ref} if words are related to insertions, deletions, and substitutions. This mode produces high-confidence SGEC training references, filtering out any corrections that may be caused by ASR errors.
\end{itemize}
Modifications are only made when edit types match, boundary positions align, and \textit{flt-hyp} and \textit{flt-ref} segments are equivalent. 
The two modes provide different trade-offs between permissiveness and precision, while both aim to ensure that the model learns grammatical corrections rather than ASR-induced discrepancies.

The full algorithm is outlined in Algorithm~\ref{algo}. It operates by iterating through alignment pairs, comparing spans and edit types, and applying the modification rules accordingly. The result is a modified GEC training reference that better reflects grammar-related corrections needed for the FLT system's output, ultimately leading to better alignment between training objectives and SGEC feedback performance. 
Although conceptually simple, the proposed alignment process yields a consistent and substantial gain in feedback precision. As shown in Section~\ref{sec:results_modify_gec}, it is the first approach that enables an end-to-end SGEC system to outperform the partial-cascaded baseline in overall feedback quality, demonstrating its strong practical impact.

\begin{algorithm}
\caption{Pseudocode for GEC Reference Modification}
\label{algo}
\begin{algorithmic}[1]
\STATE \textbf{Input:} FLT hypothesis $H$, FLT reference $R$, GEC reference $C$
\STATE \textbf{Output:} Modified \textit{mild} and \textit{strong} GEC references

\STATE $AL1 \gets \textsc{Align}(C, H)$ \COMMENT{Get word alignments}
\STATE $AL2 \gets \textsc{Align}(C, R)$

\STATE $l_{mild}, l_{strong} \gets [\ ], [\ ]$

\FOR{each $align \in AL1$}
    \STATE $(type, c_{st}, c_{end}, h_{st}, h_{end}) \gets align$
    
    \IF{$type = \text{`equal'}$}
        \STATE $l_{mild}.\textsc{Add}(C[c_{st}:c_{end}])$
        \STATE $l_{strong}.\textsc{Add}(C[c_{st}:c_{end}])$
        \STATE \textbf{continue}
    \ENDIF
    
    \STATE $f_{mild}, f_{strong} \gets \textsc{False}, \textsc{False}$
    \IF{$type \in \{\text{`ins'}, \text{`del'}\}$}        \STATE $l_{mild}.\textsc{Add}(C[c_{st}:c_{end}])$ 
        \STATE $f_{mild} \gets \textsc{True}$
     \ENDIF

    \FOR{each $alt \in AL2$}
        \IF{$type=alt.type$}
            \IF{$H[h_{st}:h_{end}] = R[alt.h_{st}:alt.h_{end}]$}
                \STATE $l_{strong}.\textsc{Add}(C[alt.c_{st}:alt.c_{end}])$ 
                \STATE $f_{strong} \gets \textsc{True}$
                \IF{$type = \text{`sub'}$}        \STATE $l_{mild}.\textsc{Add}(C[alt.c_{st}:alt.c_{end}])$ 
                \STATE $f_{mild} \gets \textsc{True}$
             \ENDIF
             \STATE \textbf{break}
            \ENDIF
        \ENDIF
    \ENDFOR

    \IF{$\neg f_{mild}$}
        \STATE $l_{mild}.\textsc{Add}(H[h_{st}:h_{end}])$ 
    \ENDIF
    \IF{$\neg f_{strong}$}
        \STATE $l_{strong}.\textsc{Add}(H[h_{st}:h_{end}])$ 
    \ENDIF
\ENDFOR
    
\RETURN \textsc{Join}($l_{mild}$), \textsc{Join}($l_{strong}$)
\end{algorithmic}
\end{algorithm}

\subsection{Confidence Estimation and Filtering for Edits}
\label{sec:confidence_desc}

Reliable feedback evaluation is essential for assessing whether an SGEC system correctly identifies learner errors. As discussed in Section~\ref{sec:align_gec}, feedback is extracted from the edits between the generated grammatically corrected transcription (\textit{gec-hyp}) and the generated fluent transcription (\textit{flt-hyp}). Since both are produced automatically by Whisper$_{\text{gec}}$ and Whisper$_{\text{flt}}$, ASR errors, disfluencies, and uncertain grammatical corrections can introduce unreliable edits.

To improve feedback reliability, we extend the confidence-based edit filtering approach of \citet{lu2022assessing} to end-to-end SGEC. Unlike their cascaded system, which estimates confidence independently from ASR, disfluency detection, and GEC modules, our approach derives edit confidence directly from the end-to-end framework. 
Specifically, the system processes each edit in the $M^2$ file, extracts the relevant words and their confidence values from both Whisper$_{\text{flt}}$ and Whisper$_{\text{gec}}$, computes the edit-confidence using the chosen aggregation mode.
Low-confidence edits are then excluded from feedback evaluation. Details in the following subsections. This two-sided, word-to-edit aggregation enables more accurate confidence estimation for E2E systems and consistently improves feedback precision, as demonstrated in Section~\ref{sec:result_confidence}.


\subsubsection{Edit-Confidence from Word-Level Scores}
Each edit produced by the $M^2$ alignment process is composed of one or more source words from the FLT transcription and one or more target words from the GEC transcription. To estimate the confidence of an edit, we associate it with two confidence sources:
\begin{itemize}
    \item FLT confidence scores: word-level confidence values generated from the Whisper$_\text{flt}$ output.
    \item GEC confidence scores: word-level confidence values from the Whisper$_\text{gec}$ output.
\end{itemize}
These scores are obtained by force-aligning the decoded transcriptions with the corresponding Whisper model and reflect the model's certainty in each word prediction.

\subsubsection{Mapping Word Confidence to edit-confidence}
To derive a confidence score for each edit, we extract the relevant word spans from both the FLT and GEC transcriptions. Depending on the type of edit (i.e., replacement, unnecessary, or missing), the corresponding tokens are mapped as follows:
\begin{itemize}
    \item Replacement (R: A $\rightarrow$ B): Use the confidence scores of the source span for A from FLT and the correction span for B from GEC.
    \item Missing (M: $\phi \rightarrow$ B): No source span in the FLT transcription. Use a fixed high confidence (e.g., 1.0) for the FLT side and the confidence of the inserted word(s) B from GEC.
    \item Unnecessary (U: A $\rightarrow \phi$): No target span in the GEC transcription. Use the confidence of the deleted word(s) A from FLT and a fixed value (such as 1.0) for GEC.
\end{itemize}
Edge cases such as apostrophes and contractions (e.g., ``I'd'') are handled through token normalization to ensure alignment consistency across both FLT and GEC sequences.

\subsubsection{edit-confidence Scoring Modes}
We define several strategies for combining the word-level scores from FLT and GEC into a single edit-confidence score. Let $a_k$ denote the $k$-th word confidence from the FLT side and $b_k$ from the GEC side. For an edit covering word indices $[i, j]$, we compute:
\begin{align}
  \text{edit}_{\mathrm{flt\text{-}avg}}
    &= \frac{1}{j - i + 1}\,\sum_{k=i}^{j} a_k\\[6pt]
  \text{edit}_{\mathrm{flt\text{-}min}}
    &= \min_{k=i,\dots,j} a_k\\[6pt]
  \text{edit}_{\mathrm{gec\text{-}avg}}
    &= \frac{1}{j - i + 1}\,\sum_{k=i}^{j} b_k\\[6pt]
 \text{edit}_{\mathrm{gec\text{-}min}}
    &= \min_{k=i,\dots,j} b_k\\[6pt]
  \text{edit}_{\mathrm{avg}}
    &= \frac{1}{2}\,\bigl(\text{edit}_{\mathrm{flt\text{-}avg}} + \text{edit}_{\mathrm{gec\text{-}avg}}\bigr)\\[6pt]
  \text{edit}_{\mathrm{min}}
    &= \min \bigl(\text{edit}_{\mathrm{flt\text{-}min}}, \text{edit}_{\mathrm{gec\text{-}min}}\bigr)
\end{align}
These scoring modes provide different trade-offs between leniency and strictness. For example, \textit{min} emphasizes the lowest-confidence token and is more conservative, whereas \textit{avg} captures overall confidence across the edit span.

\subsubsection{Filtering by edit-confidence}
After computing the edit-confidence, a threshold can be applied to filter out low-confidence edits. These filtered edits are excluded from feedback evaluation metrics, reducing noise from uncertain corrections and focusing evaluation on more reliable edits. This threshold-based filtering can be tuned according to the application. Since edit-confidence filtering is applied after SGEC inference, it does not alter the generated transcriptions or their WER. It only affects feedback evaluation by filtering low-confidence edits.


\section{Experimental Setup}
\label{sec:exp_setup}
\subsection{Datasets}
\label{sec:dataset}
This work leverages both spoken and written GEC corpora for training and evaluating GEC systems. The written GEC corpora, EFCAMDAT~\citep{geertzen2013automatic} and BEA-2019 data~\citep{bryant2019bea}, are used to train the text-based GEC model in the cascaded and partial-cascaded  GEC systems. The spoken GEC data include both the in-house Linguaskill corpus \citep{ludlow2020official}, as well as the publicly available Speak \& Improve (S\&I) Corpus \citep{knill2024speakimprovecorpus,knill2025introducing}.  The labeled and unlabeled portions of Linguaskill are used to train end-to-end SGEC models. The Linguaskill Dev set is used for hyperparameter tuning, while the Linguaskill Test set and the S\&I Dev set serve as primary evaluation sets.
While the S\&I Eval set was only released recently (March 2025) and is not used for development in this work, we report results of our best systems on it for completeness, as doing so facilitates future benchmarking and comparison by the broader research community.

\noindent \textbf{Linguaskill:} The Linguaskill corpus contains candidate responses to the Speaking module of the Linguaskill tests for L2 learners of English, provided by Cambridge University Press \& Assessment\footnote{The Linguaskill test has been renamed the Cambridge English Skills Test since this data was collected \url{https://www.cambridgeenglish.org/exams-and-tests/cambridge-english-skills-test/}}~\citep{ludlow2020official}.
The data is gender-balanced including speakers with approximately 30 different first languages (L1s), with proficiency levels ranging from A2 to C according to the Common European Framework of Reference (CEFR)~\citep{cefr2001}. It comprises two parts: a manually annotated subset (LNG$_\text{lbl}$) labeled with disfluencies and grammatical error corrections, and a much larger unlabeled subset (LNG$_\text{unl}$). Since responses can be up to 60 seconds, LNG$_\text{lbl}$ are segmented into sentence-like utterances using automatic time alignment based on manually marked phrase boundaries. Pseudo transcriptions for LNG$_\text{unl}$ are generated using the pipeline described in Section~\ref{sec:pseudo-label}. V1 and V2 correspond to pipelines using the Whisper \texttt{small.en} and \texttt{large-v2} model variants, respectively.

\noindent \textbf{S\&I:} The Speak \& Improve (S\&I) Corpus 2025~\citep{knill2024speakimprovecorpus,knill2025introducing} is a publicly available dataset of L2 learner speech, collected via the S\&I version 1 platform between 2019 and 2024~\citep{nicholls23_interspeech}. It provides diverse learner audio recordings, manual transcriptions, disfluency annotations, grammatically corrected transcripts, and associated CEFR proficiency scores from A2 to C. The corpus is created to support research in spoken language assessment and feedback. The Train and Dev sets were released in December 2024, followed by the Eval set in March 2025. In this work, the Dev set is used for primary evaluation, and results on the recently released Eval set of the best models are also reported to support future benchmarking.

\noindent \textbf{Switchboard Reannotated Dataset:} The Switchboard Reannotated Data-set\footnote{\label{note1}\url{https://github.com/vickyzayats/switchboard_corrected_reannotated}} is a refined version of the widely used Switchboard I~\citep{godfrey1992switchboard} and Release 2~\citep{godfrey1993switchboard} corpora. It aligns the corrected MsState transcriptions~\citep{deshmukh1998resegmentation}, which offer improved word accuracy and time alignment, with detailed disfluency annotations from earlier releases~\citep{marcus1999treebank}. These annotations cover common phenomena in conversational speech, such as restart, repetition, and repair, making the dataset a valuable resource for studying spontaneous spoken language.
The dataset includes predefined train, dev, and test splits~\citep{charniak2001edit}, and is openly available\footnotemark[\getrefnumber{note1}]. In this work, the reannotated Switchboard corpus is used to train the baseline Disfluency Detection (DD), which is described in Section~\ref{sec:model_setup}.

\noindent \textbf{BEA-2019:} The BEA-2019 dataset\footnote{\url{https://www.cl.cam.ac.uk/research/nl/bea2019st/}} is a large collection of English learner texts with annotations for grammatical, lexical, and spelling errors~\citep{bryant2019bea}, introduced as part of the BEA-2019 Shared Task on Grammatical Error Correction (GEC). It includes data from several well-known sources such as the First Certificate in English (FCE), Write \& Improve + LOCNESS, Lang-8 Corpus of Learner English, and the National University of Singapore Corpus of Learner English (NUCLE). The corpus spans a wide range of learner proficiency levels and L1 backgrounds, making it a valuable resource for training and evaluating GEC systems. In this work, the BEA-2019 training set is used to train the baseline GEC model, while the development set is composed of the WI+LOCNESS dev set and the FCE dev set. To ensure consistency with spoken language transcriptions, both punctuation and capitalization are removed during preprocessing. Orthographic and spelling errors are removed, as these phenomena do not typically arise in speech and are not targeted in spoken GEC correction.

Further details about the data sets can be found in Table~\ref{tab:data_stats}.

\begin{table}[!t]
\footnotesize
    \centering
    \begin{tabular}{@{ }l|@{ }l@{ }|c|c|c|c|c@{ }}
    \toprule
    & Corpus &  Split & Hours & Speakers & Utts/Sents & Words \\
    \midrule
    \multirow{7}*{\rotatebox{90}{Spoken}} & \multirow{3}*{LNG$_\text{lbl}$} & Train & 77.6 & 1,908 & 34.8K & 502K \\ 
    & & Dev & 7.8 & 176 & 3.3K & 49K \\ 
    & & Test & 11.0 & 271 & 4.6K & 69K \\ 
    \cmidrule{2-7}
    & LNG$_\text{unl}$ V1 & Train & 2521.6 & - & 708,613 & 15M \\
    & LNG$_\text{unl}$ V2 & Train & 2585.6 & - & 716.9K & 15M \\ 
    \cmidrule{2-7}
    & \multirow{2}*{S\&I} & Dev & 20.8  & - & 6.9K & 105K \\ 
    & & Eval & 20.4 & - & 6.9K & 102K \\ 
    \midrule
    \multirow{5}{*}{\rotatebox{90}{Written}} 
  & \multirowcell{2}{Switchboard} 
  & \multirowcell{2}{Train\\Dev} 
  & \multirowcell{2}{--\\--} 
  & \multirowcell{2}{1,866\\86} 
  & \multirowcell{2}{170.1K\\9.8K}
  & \multirowcell{2}{1.2M\\82K}  \\
  &&&&&& \\
\cmidrule{2-7}
  & \multirowcell{3}{EFCAMDAT\\+BEA-2019} 
  & \multirowcell{3}{Train\\Dev} 
  & \multirowcell{3}{--\\--} 
  & \multirowcell{3}{--\\--} 
  & \multirowcell{3}{2.5M\\25.5K}
  & \multirowcell{3}{28.9M\\293K} \\
    &&&&&& \\
    &&&&&& \\
    \bottomrule
    \end{tabular}
    \caption{Statistics of datasets.}
    \label{tab:data_stats}
\end{table}

\subsection{Model Details}
\label{sec:model_setup}


This work builds on Whisper foundation models to train end-to-end SGEC systems. We develop two key variants:
\begin{itemize}
    \item Whisper$_\text{flt}$, trained with fluent reference transcriptions to remove disfluencies.
    \item Whisper$_\text{gec}$, trained with grammatically corrected transcriptions to directly generate corrected outputs from speech.
\end{itemize}
Two Whisper model sizes are used as backbones: \texttt{small.en} and \texttt{large-v2}. Both fine-tuned on the Linguaskill training set using either fluent or GEC references as targets, the GEC references include manual references and pseudo-labels generated through the pipeline described in Section~\ref{sec:pseudo-label}. Initial experiments are conducted using both model sizes to explore performance trends and validate the proposed methods. In later stages, we focus on the \texttt{large-v2} model to maximize performance, particularly for feedback generation tasks, where larger model capacity is beneficial.
The \texttt{small.en} model is fine-tuned for 30,000 steps with a batch size of 5. The \texttt{large-v2} model is fine-tuned for 2 epochs with a batch size of 1 and a gradient accumulation step of 8. The small batch size is chosen due to GPU memory constraints. In both cases, the learning rate is initialized to 1e-6 and linearly decayed. During inference, decoding is performed using beam search with a beam width of 5.


To support disfluency removal in the cascaded baseline, a text-based DD model is employed. This model is initialized from \emph{bert-base-uncased}\footnote{\label{note3}\url{https://huggingface.co/google-bert/bert-base-uncased}} and fine-tuned for binary token classification, i.e., fluent or disfluent tokens. A lightweight classification head is added to the BERT encoder, consisting of a linear projection to 128 dimensions, a dropout layer with a probability of  0.2, and a final linear layer for two-class prediction. The model is trained on the Switchboard Reannotated Dataset (see Section~\ref{sec:dataset}) for 3 epochs with a learning rate of 5e-6. This DD module is integrated into the cascaded SGEC pipeline, preceding the GEC component described below.


To compare with the end-to-end systems, a text-based GEC model is used in both the cascaded and partial-cascaded  baselines. The GEC model is initialized  from the BART architecture~\citep{bart2020}, specifically the \emph{bart-base}\footnote{\label{note4}\url{https://huggingface.co/facebook/bart-base}}  from the HuggingFace Transformers library~\citep{huggingface}. It is fine-tuned on a combination of the BEA-2019~\citep{bryant2019bea} training set and EFCAMDAT~\citep{geertzen2013automatic} (Section~\ref{sec:dataset}). 
The model was trained for 19 epochs with maximum sequence length 256, batch size 16, gradient accumulation step 4, and learning rate 2e-6. It was further fine-tuned on the Linguaskill data for 5 epochs with the encoder frozen, and the learning rate reduced to 1e-5. This GEC model serves as a strong and consistent baseline for benchmarking against end-to-end SGEC systems presented in this work.



\subsection{Evaluation Metrics}
\label{sec:metrics}
Evaluating SGEC systems involves measuring both transcription accuracy and feedback quality. In this work, we adopt complementary metrics for two aims:\\
\textbf{Word Error Rate (WER)}: WER is used to assess transcription quality, aligning with prior studies~\citep{banno2024towards, lu2022assessing}. Though Translation Edit Rate (TER) has also been used, previous work shows both metrics produce similar trends. Therefore, WER is used as the primary transcription evaluation metric for simplicity and interpretability.\\
\textbf{Precision, Recall, and F$_{0.5}$}: To evaluate feedback generation, we follow established practice in written GEC evaluation and apply the MaxMatch ($M^2$) framework~\citep{dahlmeier2012better}, which identifies phrase-level edits between a pair of text sequences. 
In this work, feedback specifically refers to edit-based grammatical feedback, represented as grammatical edits extracted using the $M^2$/ERRANT framework rather than natural-language feedback or pedagogical explanations. Therefore, unlike WER, which measures transcription quality, $M^2$-based metrics evaluate whether the system identifies the correct error locations and corrections, providing an edit-level proxy for feedback quality.
Feedback quality is assessed by comparing: (a) reference edits, generated from manual GEC transcription and manual fluent transcription, (b) predicted edits, generated by comparing model-generated GEC and fluent transcriptions. Based on these edits, ERRANT~\citep{bryant2017automatic} is then used to compute Precision, Recall, and F$_{0.5}$ scores. The F$_{0.5}$ score (rather than F$_{1}$) is chosen in line with the CoNLL-2014 Shared Task~\citep{ng2014conll} to prioritize Precision over Recall, reflecting the importance of high-confidence feedback in learner-facing applications where trust and reliability are critical.

\section{Results and Discussions}
\label{sec:results}



\subsection{Baseline Comparison}
\label{sec:results_baseline}
This section re-establishes and extends the baseline comparison first reported in \citet{banno2024towards}. While the earlier study evaluated small-scale Whisper models and focused solely on Linguaskill, the present work employs both Whisper small.en and large-v2 models and evaluates them on the proprietary Linguaskill and the public S\&I corpus. This broader setup provides a stronger and more general benchmark for subsequent analyses.
All models are trained on the manually annotated LNG$_\text{lbl}$ training set and evaluated on both the in-domain LNG$_\text{lbl}$ test set (hereafter denoted as LNG) and the out-of-domain S\&I Dev set. 
Detailed training and configuration setups for each system are described in Section~\ref{sec:exp_setup}. 

\begin{table}[!t]
    \centering
    \begin{tabular}{l|cc|cc}
    \toprule
    \multirow{2}*{Model} & \multicolumn{2}{c|}{small.en} & \multicolumn{2}{c}{large-v2}\\
    & LNG & S\&I & LNG & S\&I \\
    \midrule
    Whisper$_{\text{dsf}}$ + DD + GEC & 13.45 & 17.35 & 12.11 & 14.49 \\
    Whisper$_{\text{flt}}$ + GEC & \textbf{13.24} & \textbf{16.91} & 11.81 & 13.99 \\
    Whisper$_{\text{gec}}$ & 13.48 & 17.76 & \textbf{11.10} &  \textbf{13.21} \\
    \bottomrule
    \end{tabular}
    \caption{WER evaluation of cascaded, partial-cascaded , and end-to-end (E2E) SGEC models. All Whisper models are trained on LNG$_{\text{lbl}}$.}
    \label{tab:results_baseline}
\end{table}

Table~\ref{tab:results_baseline} presents WER results across the three system architectures. Using the \texttt{small.en} model, the partial-cascaded  system (Whisper$_{\text{flt}}$ + GEC) achieves notable gains over the cascaded system (Whisper$_{\text{dsf}}$ + DD + GEC), reducing WER by 0.21\% absolute on LNG and 0.44\% on S\&I. This suggests that integrating ASR and disfluency detection reduces ASR error propagation, leading to better downstream GEC performance.
However, the E2E model (Whisper$_{\text{gec}}$) with \texttt{small.en} underperforms both baselines. This suggests that, despite architectural advantages, the E2E model struggles to generalize effectively when trained on limited LNG data. Case analysis reveals under-correction and inconsistent edit behavior, likely due to insufficient training data. 
To address this limitation, we introduce the pseudo-labeling strategy described in Section~\ref{sec:pseudo-label}, with results presented in Section~\ref{sec:results_pseudo}.
In contrast, the \texttt{large-v2} model demonstrates the benefit of scale. The partial-cascaded  \texttt{large-v2} system performs similarly to its \texttt{small.en} counterpart, achieving a 2.5\% relative WER improvement on LNG and 3.5\% on S\&I compared to the cascaded baseline. Notably, the E2E model with \texttt{large-v2} surpasses both baselines by a substantial margin, achieving a 6.9\% relative WER reduction on LNG and 5.6\% on S\&I compared to the partial-cascaded  system, and 8.3\% and on LNG and 8.8\% on S\&I over the cascaded system. These results demonstrate the strong potential of larger models, showing their larger capacity. Even when trained on a relatively small manually labeled dataset, the large-scale E2E model can outperform traditional pipelines, offering a simpler and more robust alternative to traditional multi-stage SGEC pipelines.

 \begin{table}[!htbp]
    \centering
    \begin{tabular}{l|l|cc}
    \toprule
    \multirow{1}*{Model} & Approach & LNG & S\&I \\
    \midrule
    Whisper$_{\text{gec}}$ & Fine-tune & \textbf{11.10} &  \textbf{13.21} \\
    \midrule
    \multirow{2}{*}{Qwen2-Audio-7B-Instruct} & Zero-shot & 26.10 & - \\
        & LoRA & 13.70 & - \\
    \midrule
    Voxtral-Small-24B-2507 & Zero-shot & 13.71 & 15.33 \\
    \bottomrule
    \end{tabular}
    \caption{WER evaluation of Whisper and multimodal LLMs.}
    \label{tab:results_baseline_audiollm}
\end{table}

Table~\ref{tab:results_baseline_audiollm} compares Whisper$_{\text{gec}}$ with two recent multimodal LLMs. Qwen2-Audio-7B-Instruct~\citep{chu2023qwenaudio,chu2024qwenaudio2} achieves relatively poor zero-shot performance (26.1\% WER), LoRA~\citep{hulora2022lora} fine-tuning substantially reduces the WER to 13.7\%, but it still underperforms Whisper$_{\text{gec}}$ by 2.6\% absolute WER on LNG. We also evaluate the more recent Voxtral-Small-24B-2507~\citep{liu2025voxtral} in the zero-shot setting. Although Voxtral considerably outperforms the zero-shot Qwen2-Audio model, it remains more than 2\% WER higher than Whisper$_{\text{gec}}$ on both LNG and S\&I. Error analysis shows that Voxtral increases insertion, deletion, and substitution errors, indicating that its lower performance is not primarily caused by over-correction but by the overall difficulty of zero-shot SGEC. Preliminary few-shot experiments yielded similar performance to the zero-shot setting. We additionally evaluated the zero-shot Voxtral outputs on the S\&I Dev set using the feedback metrics adopted in this work. Voxtral achieves an F$_{0.5}$ score of 23.69, substantially below the partial-cascaded baseline (43.10).
These results suggest that current zero-shot multimodal LLMs remain less effective for both learner speech correction and grammatical feedback generation. We therefore focus the remainder of this work on comparing SGEC architectures under a common Whisper backbone.


\subsection{Effect of Pseudo-labeling}
\label{sec:results_pseudo}

In the previous section, we observe that the \texttt{small.en} variant of Whisper$_{\text{gec}}$ underperforms the baseline systems, likely due to insufficient labeled training data. We therefore explore whether pseudo-labeled data can improve model performance by significantly expanding the training corpus.

As described in Section~\ref{sec:pseudo-label}, we construct two versions of the pseudo-labeled dataset. LNG$_{\text{unl}}$ V1 is generated using the \texttt{small.en} variants of Whisper$_{\text{dsf}}$ and Whisper$_{\text{flt}}$, while LNG$_{\text{unl}}$ V2 uses the corresponding \texttt{large-v2} models. Both datasets contain approximately 2,500 hours of pseudo-labeled learner speech.
For the \texttt{small.en} Whisper$_{\text{gec}}$ model, pseudo-labeling is effective. As shown in Table~\ref{tab:results_pseudo}, pre-training on LNG$_{\text{unl}}$ V1 followed by fine-tuning on LNG$_{\text{lbl}}$ reduces the WER from 13.48\% to 12.72\% on LNG and from 17.76\% to 16.84\% on S\&I, outperforming the partial-cascaded baseline on both datasets. These results demonstrate that pseudo-labeled data effectively alleviates data scarcity for smaller models.
For the \texttt{large-v2} Whisper$_{\text{gec}}$ model, pseudo-labeling is considerably less effective. Using LNG$_{\text{unl}}$ V2 provides better performance at the pre-training stage, however, after subsequent fine-tuning on the manually annotated LNG$_{\text{lbl}}$ set, both pseudo-labeling settings converge close to the performance of training on the labeled data alone (11.00\% vs. 11.10\% on LNG, and 13.20\% vs. 13.21\% on S\&I).

\begin{table}[!t]
    \centering
    \begin{tabular}{c|l|ll|ll}
    \toprule
    Model & Training Set & \multicolumn{2}{c|}{small.en} & \multicolumn{2}{c}{large-v2} \\
    & & LNG & S\&I & LNG & S\&I \\
    \midrule
    \multicolumn{2}{c|}{Whisper$_{\text{flt}}$ + GEC} & 13.24 & 16.91 & 11.81 & 13.99 \\
    \cmidrule{1-6}
    \multirow{5}*{Whisper$_{\text{gec}}$}  & LNG$_{\text{lbl}}$ & 13.48 & 17.76 & 11.10$^\dagger$ &  13.21$^\dagger$ \\
     & LNG$_{\text{unl}}$ V1 & 14.16 & 18.11 & 12.93 & 15.92 \\
       & \hspace{2.5mm} + LNG$_{\text{lbl}}$ & \textbf{12.72} & \textbf{16.84} & \textbf{11.10$^\dagger$} & \textbf{13.93} \\ 
     & LNG$_{\text{unl}}$ V2 & - & - & 12.00 & 14.12 \\
       & \hspace{2.5mm} + LNG$_{\text{lbl}}$ & - & - & \textbf{11.00$^\dagger$} & \textbf{13.20$^\dagger$} \\ 
    \bottomrule
    \end{tabular}
    \caption{Evaluation (WER) of Whisper$_{\text{gec}}$ performance with pseudo-labeled data on LNG$_{\text{lbl}}$ test and S\&I dev sets. LNG$_{\text{unl}}$ V1 and V2 refer to the pseudo-labeled data generated based on the small.en and large-v2 models, respectively. $\dagger$ indicates a statistically significant improvement over the partial-cascaded  system ($p < 0.001$).}
    \label{tab:results_pseudo}
\end{table}


These results suggest two key insights. First, pseudo-labeled data is particularly valuable for small models, which are more sensitive to data scarcity. Second, for larger models, the benefit is limited once high-quality manual annotations are introduced, as the model can already generalize effectively with fewer training examples of higher quality. Since models based on \texttt{large-v2} consistently show stronger performance, all subsequent analyses focus on the Whisper \texttt{large-v2} variant.



\subsection{Impact of Prompt-conditioned SGEC}
\label{sec:results_prompt}
In this section, we evaluate the prompted SGEC framework proposed in Section~\ref{sec:prompt}. Specifically, we compare the prompted model (Whisper$_{\text{gec+pt}}$) with the non-prompted end-to-end model (Whisper$_{\text{gec}}$) and the partial-cascaded system (Whisper$_{\text{flt}}$ + GEC).
Following Section~\ref{sec:prompt}, fluent transcriptions generated by Whisper$_{\text{flt}}$ are used as prompts during both training and inference. All models are built on the Whisper \texttt{large-v2} backbone, which showed the strongest overall performance in the previous section.

As shown in Table~\ref{tab:results_prompt_wer}, Whisper$_\text{gec+pt}$ slightly outperforms the non-prompted Whisper$_\text{gec}$ on both evaluation sets. It achieves a WER of 11.08\% on LNG and 13.09\% on S\&I, compared to 11.10\% and 13.21\%, respectively, for the non-prompted model. This Whisper$_\text{gec+pt}$ significantly outperforms the partial-cascaded  baseline, which achieves WERs of 11.81\% on LNG and 13.99\% on S\&I. The improvements of both E2E models over the partial-cascaded  system are statistically significant ($p < 0.001$). These results show that prompting with fluent transcriptions provides useful contextual information, helping refine grammatical corrections even in a strong, large-scale E2E model.


\begin{table}[!t]
\small
    \centering
    \begin{tabular}{@{ }l@{ }|@{ }c@{ }|p{8.5mm}p{8.5mm}|p{7.6mm}p{7.6mm}p{7.6mm}|p{7.6mm}p{7.6mm}p{7.6mm}}
    \toprule
    & & LNG & S\&I & \multicolumn{3}{c|}{LNG} & \multicolumn{3}{c}{S\&I} \\
    Model Name & Prompt & WER & WER & P & R & F\textsubscript{0.5} & P & R & F\textsubscript{0.5} \\
    \midrule
    \multicolumn{2}{c|}{Whisper$_\text{flt}$ + GEC} & 11.81 & 13.99  & 46.60 & 26.61 & 40.51  & 49.43 & 28.51 & 43.10 \\
    \cmidrule{1-10}
    Whisper$_\text{gec}$ & \ding{55} & 11.10$^\dagger$ & 13.21$^\dagger$ & 40.38 & 31.91 & 38.24 & 41.87 & 33.06 & 39.75 \\
    Whisper$_\text{gec+pt}$ & \ding{51} &  11.08$^\dagger$ & 13.09$^\dagger$ & 43.92 & 32.63 & 41.08 & 45.56 & 33.88 & 42.62\\
    \bottomrule
    \end{tabular}
    \caption{SGEC performance of E2E models with and without prompts, using Whisper large-v2. $\dagger$ denotes statistically significant improvement over the partial-cascaded  system ($p < 0.001$).}
    \label{tab:results_prompt_wer}
\end{table}

Although gains in SGEC accuracy are modest, the effect of prompting is more obvious in feedback generation. 
The right part of Table~\ref{tab:results_prompt_wer} reports feedback performance in terms of Precision (P), Recall (R), and F\textsubscript{0.5} (Section~\ref{sec:metrics}).
Whisper$_\text{gec+pt}$ outperforms  Whisper$_\text{gec}$ across all metrics, with a particularly notable increase in Precision, rising by 3.54 on LNG and 3.69 on S\&I. F\textsubscript{0.5} scores also increase, indicating that prompting with fluent transcriptions enables the model to better identify and express corrections aligned with learner errors. Importantly, Whisper$_\text{gec+pt}$ achieves F\textsubscript{0.5} scores on par with the partial-cascaded  baseline (Whisper$_\text{flt}$ + GEC) on both datasets, showing that a well-trained end-to-end model can match or surpass the effectiveness of pipeline-based systems, even in the more challenging task of feedback generation.

\subsection{Improvements from Aligned GEC References}
\label{sec:results_modify_gec}
Section~\ref{sec:align_gec} introduces the motivation and algorithms behind GEC reference alignment, designed to improve SGEC feedback by reducing the impact of ASR-related errors. 
Here, we evaluate the effectiveness of this approach in feedback performance, using the best-performing E2E model from previous experiments, i.e. Whisper$_\text{gec+pt}$ \texttt{large-v2}. 
When trained with the original unaligned (Orig) GEC references, this model achieves F\textsubscript{0.5} scores of 41.08 on LNG and 42.62 on S\&I. Three different versions of GEC references are compared: Orig (unaligned), Mild, and Strong. These reference types differ in the extent to which they align with the fluent transcription structure, as described in Section~\ref{sec:align_gec}. The training setup remains unchanged across experiments, with only the reference transcriptions being varied.

Results are presented in Table~\ref{tab:edit_confidence}. As expected, with modified GEC references, the models achieve higher Precision and lower Recall. This trade-off is beneficial in feedback generation, where high precision edits are generally preferred to avoid overcorrection. In terms of F\textsubscript{0.5}, which balances both metrics with an emphasis on Precision, both aligned references outperform Orig GEC. Mild GEC achieves an F\textsubscript{0.5} improvement of 1.06 on LNG and 1.67 on S\&I over the Orig GEC. Strong GEC yields the highest overall scores, with F\textsubscript{0.5} reaching 43.69 on LNG and 45.96 on S\&I, corresponding to a relative improvement of 6.4\% and 7.8\%, respectively. 
These results highlight the effectiveness of the proposed reference alignment approach in enhancing feedback quality. The Mild reference alignment strategy offers a balanced outcome, i.e. gains in F\textsubscript{0.5} while preserving WER (Orig: 11.08\% WER on LNG, 13.09\% on S\&I; Mild: 11.07\% WER on LNG, 13.09\% on S\&I), making it a practical choice when both transcription accuracy and feedback reliability are important. In contrast, the Strong alignment strategy delivers the highest F\textsubscript{0.5} scores on both datasets, demonstrating substantial improvements in feedback precision. This gain, however, comes at the cost of a slight increase in WER (11.52\% on LNG, 13.32\% on S\&I), reflecting a trade-off between transcription accuracy and feedback performance.
Importantly, both Mild and Strong alignment approaches enable the end-to-end SGEC model to surpass the partial-cascaded baseline in overall feedback quality in terms of F\textsubscript{0.5}, highlighting its significant practical impact despite its conceptual simplicity.

\subsection{Feedback with Confidence}
\label{sec:result_confidence}
This section evaluates the edit-confidence filtering strategy proposed in Section~\ref{sec:confidence_desc}. Since filtering is applied after SGEC decoding, it does not affect the generated transcriptions or their WER, but only the evaluation of grammatical feedback. For each dataset, the edit-confidence threshold was determined by sweeping values from 0.00 to 1.00 in increments of 0.01 and selecting the threshold that maximized F$_{0.5}$.

\begin{table}[!t]
\small
    \centering
    \begin{tabular}{@{}l@{ }|@{ }l@{ }|@{ }l@{ }|@{ }lll@{ }|@{ }l@{ }|@{ }lll@{}}
    \toprule
    \multirow{1}{*}{Model} & \multirow{1}{*}{Train} & \multicolumn{4}{c@{ }|@{ }}{LNG} & \multicolumn{4}{c}{S\&I} \\
      Name & Ref & WER & P & R & F\textsubscript{0.5} & WER&  P & R & F\textsubscript{0.5}  \\
    \midrule
    \multicolumn{2}{@{ }c|@{ }}{Whs$_\text{flt}$ + GEC} &  11.81 & 46.60 & 26.61 & 40.51 & 13.99 & 49.43 & 28.51 & 43.10 \\
    \cmidrule{1-10}
    \multirow{2}{*}{Whs$_\text{gec}$} &  Orig & \multirow{2}{*}{11.10} & 40.38 & 31.91 & 38.24 & \multirow{2}{*}{13.21} & 41.87 & 33.06 & 39.75 \\
    &  \hspace{2mm}+ $\text{edit}_{\mathrm{avg}}$ & & 51.84 & 24.79 & 42.55 & & 53.17 & 26.45 & 44.23\\
    \cmidrule{1-10}
    \multirow{6}{*}{Whs$_\text{gec+pt}$} & Orig & \multirow{2}{*}{11.08} & 43.92 & 32.63 & 41.08  & \multirow{2}{*}{13.09} & 45.56 & 33.88 & 42.62 \\
    &  \hspace{2mm}+ $\text{edit}_{\mathrm{avg}}$ & & 51.10 & 27.58 & 43.65 & & 54.08 & 28.20 & 45.69 \\
    \cmidrule{2-10}
    & Mild & \multirow{2}{*}{11.07} & 46.96 & 29.88 & 42.14 & \multirow{2}{*}{13.09} & 49.26 & 31.55 & 44.29\\
    &  \hspace{2mm}+ $\text{edit}_{\mathrm{avg}}$ & & 51.23 & 27.87 & 43.87 & & 54.89 & 27.98 & 46.04\\
    \cmidrule{2-10}
    & Strong & \multirow{2}{*}{11.52} & 52.20 & 26.45 & 43.69  &  \multirow{2}{*}{13.32} & 55.03 & 27.71 & 45.96\\
    &  \hspace{2mm}+ $\text{edit}_{\mathrm{avg}}$ & & 55.65 & 25.06 & 44.73 && 58.03 & 26.61 & 46.94 \\
    \bottomrule
    \end{tabular}
    \caption{Feedback performance (Precision, Recall, F$_{0.5}$) of various end-to-end SGEC models using Whisper large-v2.}
    \label{tab:edit_confidence}
\end{table}


We compare six edit scoring methods across four models, evaluated both on LNG test and S\&I Dev sets. The results show consistent trends: filtering increases Precision, slightly reduces Recall, and ultimately improves F\textsubscript{0.5}. Taking Whisper$_\text{gec+pt}$ (Orig) on LNG as an example, using FLT-based confidence yields notable improvements in F\textsubscript{0.5} over the baseline without filtering (F\textsubscript{0.5}=41.08), with \textit{flt-avg} reaching 42.85 and \textit{flt-min} achieving 43.06. GEC-only scores achieve smaller gains (42.40 with \textit{gec-avg}, 42.26 with \textit{gec-min}). The best results are obtained by combining both confidence sources: 43.65 with \textit{avg} and 43.09 with \textit{min}. These improvements are comparable to those obtained using Strong Reference alignment (Section~\ref{sec:results_modify_gec}) and are complementary to it, indicating that edit-confidence filtering provides an effective and complementary way to enhance SGEC feedback evaluation. Consistent trends were observed on the S\&I set and across other model variants.

Table~\ref{tab:edit_confidence} summarizes SGEC results with and without  \textit{avg} edit-confidence filtering.  Across all settings, Whisper$_\text{gec}$ and Whisper$_\text{gec+pt}$, and with Orig, Mild or Strong GEC reference, edit-confidence filtering consistently improves F\textsubscript{0.5}. 
The best-performing configuration on LNG is Whisper$_\text{gec+pt}$ with the Strong GEC reference and \textit{edit$_\text{avg}$}, achieving an F$_\text{0.5}$ of 44.73, and the best configuration for S\&I is the same 
setup, achieving an F$_\text{0.5}$ of 46.94. This reflects a relative improvement of 2.4\% and 2.1\%  over the unfiltered baseline (43.69 and 45.96, respectively). 
To our knowledge, this is the first integration of reference alignment and edit-confidence filtering within an end-to-end SGEC framework, jointly enhancing feedback reliability beyond what either technique achieves alone.

\begin{table}[!t]
\centering
\small
\begin{tabular}{l|cc|cc}
\toprule
& \multicolumn{2}{c|}{LNG} & \multicolumn{2}{c}{S\&I} \\
Setting & th & F$\textsubscript{0.5}$ & th & F$\textsubscript{0.5}$ \\
\midrule
Dataset-specific & 0.67 & \textbf{44.73} & 0.65 & \textbf{46.94} \\
Cross-dataset & 0.65 & 44.57 & 0.67 & 46.86 \\
\bottomrule
\end{tabular}
\caption{Threshold robustness of edit-confidence filtering for the best-performing configuration (Whisper$_{\text{gec+pt}}$ with Strong reference alignment and \textit{edit}$_{\mathrm{avg}}$). The transferred threshold uses the optimal threshold from the other dataset.}
\label{tab:threshold_transfer}
\end{table}

The optimal thresholds are similar across the two datasets, differing by only 0.02 for the best-performing configuration (Whisper$_{\text{gec+pt}}$ with Strong reference alignment and \textit{edit}$_{\mathrm{avg}}$). To assess the robustness of threshold selection, we additionally applied the threshold optimized on LNG to S\&I and vice versa. As shown in Table~\ref{tab:threshold_transfer}, transferring the threshold changes F$\textsubscript{0.5}$ by at most 0.16 points. Similar trends are observed for all other configurations, indicating that edit-confidence filtering is robust to the precise threshold choice and generalizes well across datasets.

\begin{figure}[!htbp]
    \centering
    \includegraphics[width=0.75\linewidth]{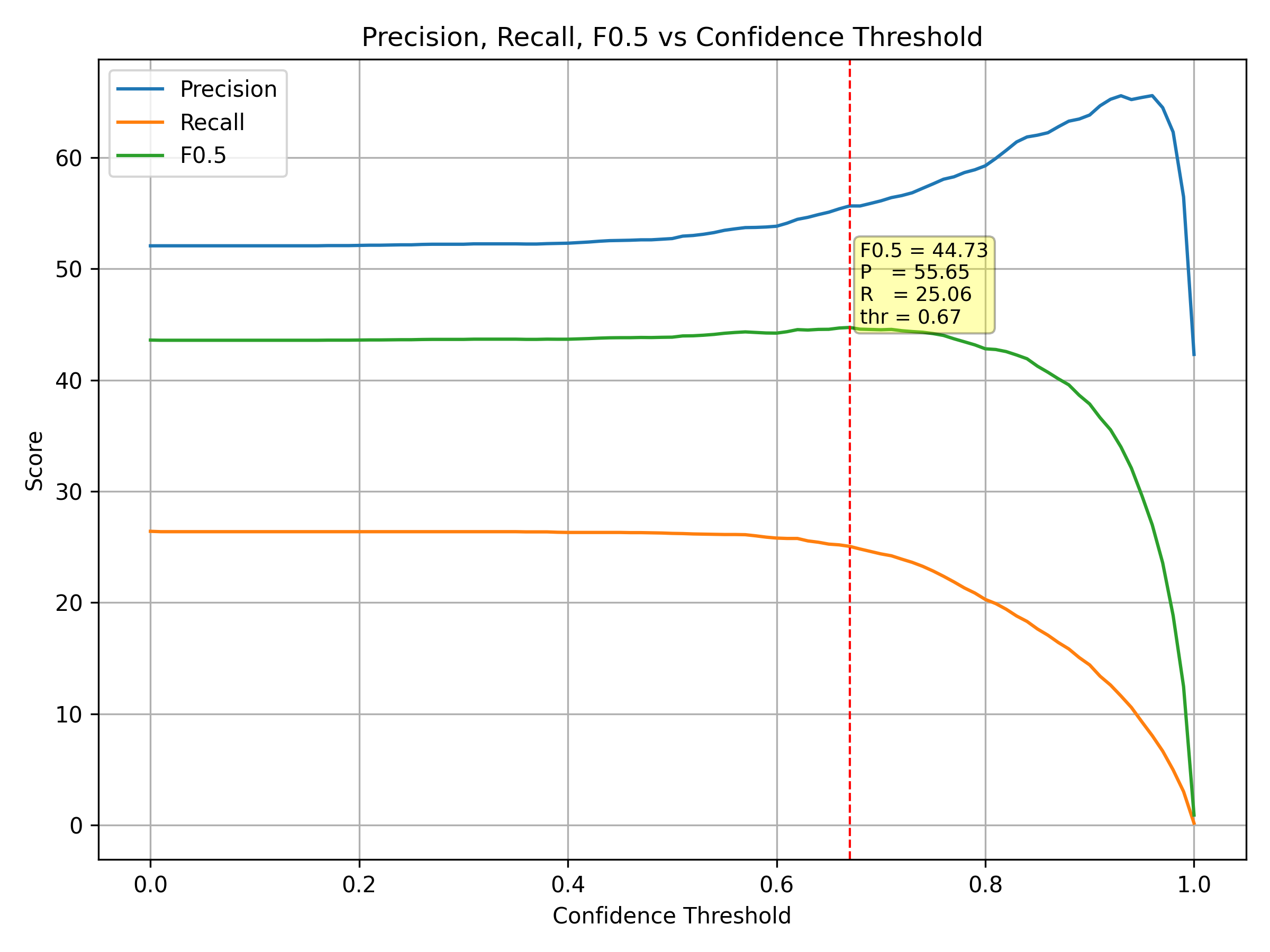}
    \caption{Precision, Recall, and F$_\text{0.5}$ scores at different edit-confidence thresholds for the Whisper$_\text{gec+pt}$ model trained with Strong GEC references.}
    \label{fig:prf_strongref_lng}
\end{figure}

To better understand the effect of threshold, Figure~\ref{fig:prf_strongref_lng} plots Precision, Recall, and F\textsubscript{0.5} against different confidence thresholds on LNG. This analysis uses the best-performing setup: Whisper$_\text{gec+pt}$ with Strong reference and \textit{edit$_\text{avg}$}. As expected, increasing the threshold improves Precision by discarding more uncertain edits, while Recall declines. F$_\text{0.5}$, which prioritizes Precision, peaks at a threshold of 0.67. This suggests that moderate filtering helps suppress noisy edits without losing too many valid corrections. 
Notably, the drop in Precision at very high thresholds is due to how default confidence scores are assigned to certain edit types. Specifically, a fixed score of 1.0 is assigned to the FLT side for the missing/deletion type of edits and to the GEC side for the unnecessary/insertion type of edits (see Section~\ref{sec:confidence_desc}). As a result, some low-quality edits remain even when the threshold is high. We experimented with other fixed values, and while the Precision and Recall curves changed slightly, the peak F$_\text{0.5}$ score remained largely unaffected.

\subsection{Case study}

\begin{table}[!t]
\small
\centering
\begin{tabular}{@{ }l@{ }|p{9.8cm}@{ }}
\toprule
\textbf{System} & \textbf{Output} \\
\midrule
\multicolumn{2}{l}{\textbf{Example 1}} \\
\midrule
FLT Reference & eating habit are very important if you will have healthy \\
GEC Reference & eating habits are very important if you want to be healthy \\
\midrule
Whs$_\text{flt}$ & is in habit are very important if you will have healthy \\
\midrule
Whs$_\text{flt}$ + GEC & having a habit is very important if you want to healthy \\
Whs$_\text{gec}$ & eating habits are very important if you want to be healthy \\
Whs$_\text{gec+pt}$ (Strong) & is in habits are very important if you want to be healthy \\
\midrule
\multicolumn{2}{l}{\textbf{Example 2}} \\
\midrule
FLT Reference & i think that the best thing of studying in a second language is that you can put in in your resume \\
GEC Reference & i think that the best thing about studying in a second language is that you can put it in your resume \\
\midrule
Whs$_\text{flt}$ & i think that the best thing of studying in a second language is that you can put in in your ration \\
\midrule
Whs$_\text{flt}$ + GEC & i think that the best thing about studying a second language is that you can put it in your ration \\
Whs$_\text{gec}$ & i think that the best thing about studying in a second language is that you can put it in your resume \\
Whs$_\text{gec+pt}$ (Strong) & i think that the best thing about studying in a second language is that you can put it in your ration \\
\bottomrule
\end{tabular}
Qualitative comparison of SGEC system outputs on selected utterances from the S\&I Dev set.
FLT and GEC denote the fluent and grammatically corrected references.
Whs$_\text{flt}$ represents a fluent ASR model, Whs$_\text{flt}$ + GEC a cascaded ASR+GEC pipeline,
Whs$_\text{gec}$ an end-to-end SGEC model, and Whs$_\text{gec+pt}$ (Strong) a prompt-conditioned variant trained with strongly aligned references.
\label{tab:example_eating}
\end{table}

To complement the quantitative results, Table~\ref{tab:example_eating} presents two representative examples from the S\&I Dev set. The examples compare the outputs of the partial-cascaded system (Whisper$_{\text{flt}}$ + GEC), the end-to-end SGEC model (Whisper$_{\text{gec}}$), and the prompted model trained with Strong reference alignment (Whisper$_{\text{gec+pt}}$ (Strong)). They illustrate the trade-off between transcription accuracy and feedback precision discussed in previous sections.

In Example 1, Whisper$_{\text{flt}}$ misrecognizes `eating' as `is in', causing the cascaded system to produce the unnatural phrase `having a habit is'. Because the downstream GEC model only operates on text, it cannot recover the speaker's intended meaning once the ASR output is corrupted. In contrast, Whisper$_{\text{gec}}$ correctly recognizes `eating habits' and generates the reference correction. The prompted model preserves the ASR hypothesis (`is in habits'), reflecting its stronger consistency with the fluent transcription used during training.

Example 2 shows a similar pattern. Whisper${_\text{flt}}$ misrecognizes `resume' as `ration', which propagates to the cascaded system. Whisper$_{\text{gec}}$ recovers the correct word directly from the speech signal, whereas Whisper$_{\text{gec+pt}}$ (Strong) retains the ASR hypothesis. Although this increases WER against the GEC reference, it avoids generating a misleading grammar edit for a recognition error, improving feedback precision.

These examples illustrate the complementary strengths of the proposed end-to-end models. Whisper$_{\text{gec}}$ is more robust to ASR errors and better preserves the speaker's intended meaning, leading to lower WER. In contrast, the prompted model trained with Strong reference alignment deliberately remains more consistent with the fluent transcription, reducing spurious grammar edits caused by ASR errors. This qualitative analysis reflects the quantitative results presented in Sections~\ref{sec:results_modify_gec} and~\ref{sec:result_confidence}, highlighting the trade-off between transcription accuracy and feedback quality.

\subsection{Results on S\&I Eval set}
To facilitate future comparisons and benchmark progress on SGEC feedback, we evaluate selected models on the recently released S\&I Eval set, which became available only in March 2025. Since the Eval set shares a similar distribution with the S\&I Dev set, we do not repeat all experimental configurations. Instead, we report results for the best-performing models identified during development: Whisper$_\text{gec+pt}$ trained with Mild and Strong reference alignments.

\begin{table}[!t]
    \centering
    \begin{tabular}{l|l|l|ccc}
    \toprule
    \multirow{1}{*}{Model} & \multirow{1}{*}{Train} & \multicolumn{4}{c}{S\&I} \\
      Name & Ref & WER & P & R & F\textsubscript{0.5}   \\
    \midrule
    \multicolumn{2}{c|}{Whs$_\text{flt}$ + GEC} &  14.37 & 48.97 & 27.25 & 42.24 \\
    \cmidrule{1-6}
    \multirow{3}{*}{Whs$_\text{gec+pt}$} & Mild & 13.44$^\dagger$  & 47.76 & 29.96 & 42.69 \\
    \cmidrule{2-6}
    & Strong & \multirow{2}*{13.72} & 53.61 & 26.29 & 44.39 \\
    & \hspace{2mm}+ $\text{edit}_{\mathrm{avg}}$ &  & 57.35 & 24.50 & \textbf{45.22} \\
    \bottomrule
    \end{tabular}
    \caption{SGEC and Feedback performance on S\&I Eval set. $\dagger$ denotes statistically significant improvement over the partial-cascaded  system ($p < 0.001$).}
    \label{tab:results_sandi_eval}
\end{table}

As shown in Table~\ref{tab:results_sandi_eval}, the Whisper$_\text{gec+pt}$ model trained with Mild references achieves the lowest WER of 13.44\%. This represents a statistically significant improvement over the partial-cascaded  baseline, which has a WER of 14.37\% ($p < 0.001$). In addition, this model also improves the feedback F\textsubscript{0.5} score from 42.24 to 42.69. This confirms that Mild reference alignment can boost feedback quality without compromising transcription accuracy.
The Strong reference configuration further improves F\textsubscript{0.5} to 44.39. When combined with edit-confidence filtering using \textit{edit\textsubscript{min}}, the model achieves the highest F\textsubscript{0.5} of 45.22, representing a substantial gain of 3.0\% over the baseline. This underscores the effectiveness of the proposed reference alignment and edit-confidence methods in enhancing SGEC feedback quality. While the Strong setup results in a slightly higher WER (13.72\%), the substantial gain in feedback precision makes it well suited for applications that prioritize accurate and targeted grammatical corrections.

\section{Conclusions}
\label{sec:conclusion}
This work addresses the challenge of SGEC in the end-to-end setup, with a particular focus on improving learner feedback.  We propose four complementary techniques: (1) a large-scale automatic pseudo-labeling pipeline that generates over 2,500 hours of SGEC training data, (2) prompt-conditioned SGEC training and inference using fluent transcriptions to enhance feedback relevance, (3) a GEC reference alignment method that reduces ASR-induced discrepancies in feedback generation, and (4) an edit-confidence estimation framework that filters unreliable feedback edits.
Experiments on both the in-house Linguaskill dataset and the public S\&I corpus demonstrate consistent improvements in end-to-end SGEC performance. Prompting with fluent references improves both transcription quality and feedback precision. Reference alignment further enhances F\textsubscript{0.5} by reducing mismatches in the training samples, consistently surpassing the partial-cascaded system for the first time in feedback generation. 
Edit-confidence filtering further removes unreliable edits, resulting in more robust feedback evaluation.
These methods enable the E2E model to outperform cascaded and partial-cascaded  baselines across key metrics.
Despite these gains, generating reliable learner feedback in the presence of ASR errors remains challenging, and there is still a trade-off between transcription accuracy and feedback precision, particularly when using more aggressive reference alignment strategies. 
Future work will focus on further improving SGEC feedback and exploring more recent multimodal LLMs for SGEC. In particular, we will investigate whether a single model can jointly generate fluent transcriptions, grammatically corrected transcriptions, and learner feedback, eliminating the need for separate fluent and SGEC models for feedback while further simplifying the overall framework.

\section*{Acknowledgments}
This work reports on research supported by Cambridge University Press \& Assessment, a department of The Chancellor, Masters, and Scholars of the University of Cambridge. Hari Krishna Vydana helped prepare the initial pseudo-labeled data. The confidence based edit pruning builds on initial work with Vyas Raina and Tom Hardman.

\section*{Declaration of generative AI and AI-assisted technologies in the manuscript preparation process}
During the preparation of this work, the author(s) used ChatGPT for reviewing grammar of the text. The author(s) reviewed and edited the output as needed and take full responsibility for the content of the published article. 

\bibliographystyle{elsarticle-harv} 
\bibliography{mybib}






\end{document}